\documentclass[
]{ceurart}

\usepackage{listings}
\usepackage{tcolorbox}
\usepackage{caption}

\newtcolorbox{promptbox}[1][]{
  colback=green!5!cyan!10,
  colframe=black,
  boxrule=0.25pt,
  left=6pt, right=6pt, top=6pt, bottom=6pt,
  title=#1
}

\begin{document}
\copyrightyear{2025}
\copyrightclause{Copyright for this paper by its authors.
  Use permitted under Creative Commons License Attribution 4.0
  International (CC BY 4.0).}

\conference{Challenge and Workshop (BC9): Large Language Models for Clinical and Biomedical NLP, International Joint Conference on Artificial Intelligence (IJCAI), August 16--22, 2025, Montreal, Canada}

\title{UETQuintet at BioCreative IX - MedHopQA:  Enhancing Biomedical QA with Selective Multi-hop Reasoning and Contextual Retrieval}

\author[1]{Quoc-An Nguyen}[
orcid=0000-0002-0605-5841,
email=annq@vnu.edu.vn,
]
\address[1]{VNU University of Engineering and Technology,  144 Xuan Thuy St, Cau Giay, Hanoi, Vietnam}

\author[1]{Thi-Minh-Thu Vu}[
orcid=0009-0001-5037-5033,
email=22028116@vnu.edu.vn,
]
\fnmark[1]

\author[1]{Bich-Dat Nguyen}[
orcid=0009-0004-3479-3827,
email=23021520@vnu.edu.vn,
]
\fnmark[1]

\author[1]{Dinh-Quang-Minh Tran}[
orcid=0009-0004-1381-1862,
email=23020554@vnu.edu.vn,
]
\fnmark[1]

\author[1]{Hoang-Quynh Le}[
    orcid=0000-0002-1778-0600,
    email=lhquynh@vnu.edu.vn,
    url=https://uet.vnu.edu.vn/~lhquynh/,
]
\cormark[1]

\cortext[1]{Corresponding author.}
\fntext[1]{These authors contributed equally.}

\begin{abstract}
Biomedical Question Answering systems play a critical role in processing complex medical queries, yet they often struggle with the intricate nature of medical data and the demand for multi-hop reasoning.
In this paper, we propose a model designed to effectively address both direct and sequential questions.
While sequential questions are decomposed into a chain of sub-questions to perform reasoning across a chain of steps, direct questions are processed directly to ensure efficiency and minimise processing overhead.
Additionally, we leverage multi-source information retrieval and in-context learning to provide rich, relevant context for generating answers.
We evaluated our model on the BioCreative IX - MedHopQA Shared Task datasets.
Our approach achieves an Exact Match score of 0.84, ranking second on the current leaderboard.
These results highlight the model’s capability to meet the challenges of Biomedical Question Answering, offering a versatile solution for advancing medical research and practice.

\end{abstract}


\maketitle

\section{Introduction}
The exponential growth of information has intensified the need for Question Answering (QA) systems, which provide accurate responses to user questions in natural language~\cite{martin2009speech,green1961baseball}.
Biomedical QA, an emerging area within QA systems, facilitates innovative solutions for accessing, comprehending, and interpreting intricate biomedical knowledge~\cite{jin2022biomedical}.

A key challenge in biomedical QA is the need for multi-hop reasoning, which involves integrating information from multiple sources—sequentially or in parallel—to answer questions. 
In particular, sequential questions require a step-by-step reasoning process, where each step builds upon the answer of the previous one.
For example, \textit{Which chromosome contains the gene most commonly associated with hereditary hemochromatosis in people of Celtic descent?} 
To answer this question, models must: 1. identify \textit{the gene} most commonly associated with hereditary hemochromatosis in individuals of Celtic descent, and 2. determine which \textit{chromosome} contains that gene. 
This two-step reasoning highlights the importance of multi-hop capabilities, as supporting information is often distributed across different sources, requiring multi-hop reasoning to reach the answer~\cite{perez2020unsupervised}.

While large language models (LLMs) have demonstrated considerable potential across domains, their performance on biomedical QA remains limited by the complexity of such multi-hop tasks.
The main reason is that biomedical questions are often highly complex, requiring domain expertise and the logical integration of information from multiple
sources~\cite{nguyen2019question}.
To address these challenges, some works have turned to Retrieval-augmented Generation (RAG), in which an information retriever selects relevant content from a large knowledge base and a language model conditions its output on that retrieved text~\cite{jeong2024improving}. 
This approach enables LLMs to incorporate specialized external knowledge alongside their pre-trained knowledge to improve performance without expensive fine-tuning~\cite{shi2025mkrag}.

In this paper, we propose a model that effectively handles both direct and sequential questions. 
While sequential questions are decomposed by a LLM, direct questions are excluded from this process to avoid potential errors introduced by the LLM. Additionally, we leverage multi-source information retrieval and in-context learning to provide relevant context for the LLM.
Our contributions are as follows:
\begin{itemize}
\item Propose a QA framework that dynamically distinguishes between direct and sequential questions to apply appropriate reasoning strategies.
\item Propose using a lightweight machine learning model to identify sequential  questions prior to decomposition, reducing unnecessary complexity.
\item Enhance answer accuracy by integrating information from multiple retrieval sources and leveraging in-context learning to provide contextual grounding.
\item Demonstrate the effectiveness of the proposed model by achieving promising results on the BioCreative IX - MedHopQA Shared Task.
\end{itemize}

\section{Related Work}
QA models have evolved significantly, transitioning from traditional retrieval-based approaches to advanced LLMs. 
Early QA systems primarily relied on classical information retrieval techniques and were often constrained to closed-domain settings~\cite{manning2009introduction}. 
With the rise of machine learning, QA systems began adopting statistical learning techniques using hand-crafted text features~\cite{rajpurkar2016squad}.
The introduction~of transformer-based architectures marked a major breakthrough in QA performance by allowing a deeper contextual understanding of input text~\cite{devlin2019bert}.
More recently, LLMs have revolutionized QA by supporting zero-shot and few-shot learning, allowing models to generalize to unseen questions without task-specific training. Furthermore, the emergence of RAG methods has addressed the limitations of fixed knowledge in LLMs by integrating external retrieval mechanisms~\cite{lewis2020retrieval}.
Multi-hop Question Answering approaches have enabled complex reasoning by incorporating external knowledge sources~\cite{chakraborty2024multi}.

In the biomedical domain, QA models must process questions that often require domain-specific knowledge and multi-step reasoning.
To enhance in-domain performance, models like BioBERT, ClinicalBERT, and PubMedBERT have been pre-trained on biomedical corpora~\cite{lee2020biobert}. 
While these domain-adapted models show potential results, they typically require extensive computational resources for pre-training and fine-tuning.
The introduction of RAG-based models marked a shift toward more efficient approaches by allowing LLMs to retrieve and integrate external information at inference time. 
An example is KG-RAG \cite{jin2022biomedical}, a biomedical QA system that integrates structured knowledge from knowledge graphs with the implicit knowledge embedded in LLMs.
However, using knowledge graphs still faces limitations regarding data completeness, predefined reasoning paths that restrict multi-step inference capabilities, and challenges in timely knowledge updating.
The development of advanced search tools has significantly enhanced the ability of models to access better performance, comprehensive, and up-to-date information sources, thereby improving their performance in tasks that required reliable external knowledge~\cite{shi2025searchrag}.

\section{Proposed Model}
\begin{figure}
  \centering
  \includegraphics[width=\linewidth]{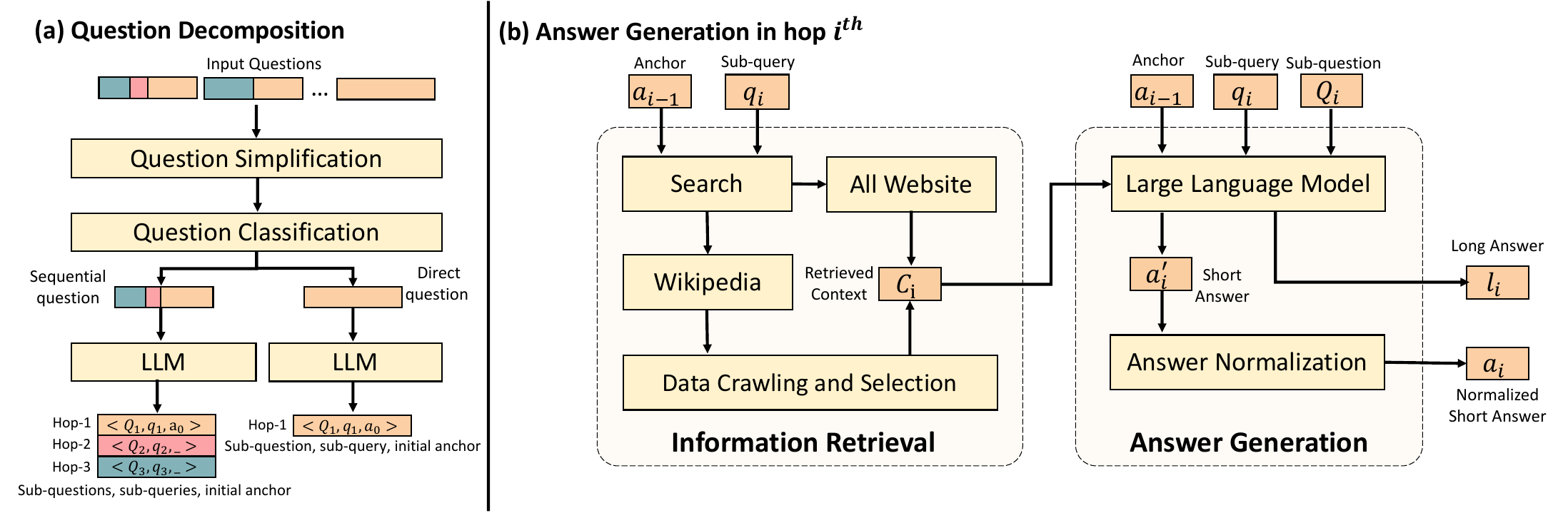}
  \caption{Overview of the proposed model. (a) Question Decomposition: The input question is simplified and classified as either a direct or a sequential question. Sequential questions are further decomposed into sub-questions. For each question, a query and an initial anchor are extracted. (b) Answer Generation at hop~$i^{th}$: The model retrieves relevant context and generates an answer using the sub-question, retrieved context and the anchor. The short answer output~$a_i$ is used as the final predicted answer if~$i$ is the last hop, otherwise, it is used as an anchor for the next hop~$i+1$.}
  \label{fig:overview}
\end{figure}

This section presents the main components of the proposed model.
The general architecture of the model is illustrated in Figure~\ref{fig:overview}.
Firstly, questions are classified and generates sub-questions, sub-queries, and an initial anchor.
In hop $i^{th}$, the model retrieves information from multiple sources and generates a predicted answer that serves as the anchor for the next hop. 
The final answer corresponds to the output of the last reasoning hop.

\subsection{Question Decomposition}
Firstly, the long questions are simplified. 
After that, we use machine learning models to classify questions into two types: sequential and direct. 
Next, sequential questions are decomposed into sub-questions using a sub-query and an initial anchor, while direct questions are handled by extracting only the sub-query and the initial anchor.

\paragraph{Question Simplification}
Some questions contain preceding explanatory sentences that may introduce distracting information for the model. 
To remove extraneous descriptive content and focus the model's attention on the core inquiry, we first identify questions that are linked to preceding declarative statements and use in-context learning to prompt the LLM to generate a simplified version.
The simplified version is required to be fewer than 50 words.

\paragraph{Question Classification}  
Some studies showed that using LLMs for question decomposition can lead to irrelevant sub-questions due to hallucination~\cite{ye2023large}.
Therefore, we propose a simple method to classify questions into two types—sequential and direct—and applied decomposition only to the sequential questions.
Firstly, we extract linguistic, structural, and embedding-based features from each question.
After that, a stacking ensemble model is constructed using Random Forest and XGBoost as base learners, with a Logistic Regression model serving as the meta-classifier to classify questions into two types. With the input feature vector $\mathbf{X}$ of input questions, we have:
\begin{equation}
\mathrm{H}(\mathbf{X}) = [\text{RF}(\mathbf{X}) \parallel \text{XGB}(\mathbf{X})]
\end{equation}
\begin{equation}
\hat{y} = \sigma(\mathbf{W} \cdot H(\mathbf{X}) + b)
\end{equation}
where \(\text{RF}(.)\) and \(\text{XGB}(.)\) denote the outputs from the Random Forest and XGBoost base learners,  
$\parallel$~is the concatenation,
$\sigma$ is the sigmoid function,
\(\mathbf{W}\) is the trainable weights, 
\(b\) is the bias, 
and \(\hat{y}\) is the prediction.

\paragraph{Question Decomposition}
If a sentence is classified as a sequential question, we decompose it into sub-questions, sub-queries, and initial anchors.
At first hop, we extract entities to highlight important keywords that need to be aware of in questions, we define it as an initial anchor.
Our approach relies on in-context learning, leveraging LLMs by providing examples within the prompt (Figure~\ref{fig:prompt1}).
On the other hand, if a sentence is a direct question, the LLM is required to extract a sub-query and the initial anchor.

\begin{center}
    \centering
    \captionof{figure}{System prompt for sequential question}
    \label{fig:prompt1}
    \fbox{
      \begin{minipage}{1.0\textwidth} 
        \scriptsize
        \texttt{<system>} \\
You are an AI expert in decomposing complex biomedical questions into a sequence of simpler, interconnected natural language sub-questions.

\textbf{Core Instructions for Decomposition:}

0.  \textbf{Faithful Reasoning via Decomposition}: Externalize reasoning process by breaking the \texttt{Original Complex Question}.

1.  \textbf{Step-by-Step Plan}: Generate a sequence of sub-questions.

2.  \textbf{Self-Contained and Searchable sub-questions}: Each sub-question MUST be self-contained in isolation (...)

3.  \textbf{Logical Dependency}: Explicitly reference prior results for logical flow, embedding context for self-containment (...)

4.  \textbf{Sub-question properties}: For each step in your decomposition:

\quad  - \quad  sub-question: State in clear, natural language, adhering to the self-contained principle and clarity of result (...)

\quad  - \quad  sub-query: Provide a distilled set of keywords from a sub-question, targeting key biomedical terms and concepts (...)

5.  \textbf{Efficiency and Necessity}: Decompose the question into the \textbf{minimum number of steps} to logically arrive at the final answer.

6.  \textbf{Final Answer Derivation}: The final sub-question's purpose is to synthesize the information gathered from all preceding, answered sub-questions to directly and comprehensively answer the \texttt{Original Complex Question}.

7.  \textbf{Output Format}: A valid JSON array of step objects.

Original Complex Question: \{ $Q$ \} 

\texttt{<user>} \\
Here are 5 examples demonstrating the desired JSON output format and decomposition logic:

\texttt{---Examples---}

    \end{minipage}
  
    }
\begin{minipage}{1.0\textwidth}
    \raggedleft
    {\footnotesize\itshape The text content has been condensed by substituting with (...)}
  \end{minipage}
\end{center}

\subsection{Answer Generation}
With a question having $n$ hops ($n=1$ if direct), the model iteratively answers each hop, using the previous answer as the anchor for the next. 
Specifically, at hop $i^{th}$, the model retrieves relevant context based on the sub-query and anchor, then using the sub-question, retrieved context, and anchor to generate an answer. 
The final short and long answers are taken from the last hop.

\paragraph{Retrieval Information}
Firstly, the search engine is used to find relevant information sources for a sub-query and an anchor. 
The input to the search engine is concatenated between sub-query  $q_i$ and anchor $a_{i-1}$.
The output is relevant information sources, including title, link, and snippet for each corresponding document.
A snippet is a short piece of text that provides a preview or excerpt of a page's content, deemed relevant to the input query by the search engine.
Wikipedia, a large open-access resource, contains a wealth of biomedical information\footnote{https://en.wikipedia.org/wiki/Wikipedia:Statistics}.
We collect Wikipedia articles and crawl data to gather additional context for the model.  
We then use the Term Frequency-Inverse Document Frequency (TF-IDF) to convert each sentence into a vector reflecting term importance. 
Cosine similarity between sentence vectors quantifies semantic similarity.
Finally, the top $m$ most similar sentences are selected as additional retrieved context for the next step.
The output of this phase is a relevant context $C_i$:
\begin{equation}
C_i = \{s_j\}^{n_s}_{j=1} \cup \{k_j\}^{m}_{j=1}
\end{equation}
where \( \{s_j\}^{n_s}_{j=1} \) is the set of snippets returned by the search engine, and \(  \{k_j\}^{m}_{j=1} \) is the set of important sentences from Wikipedia.

\paragraph{Answer Generation}
The input of this phase contains sub-question $Q_i$, sub-query $q_i$, anchor $a_{i-1}$ and relevant context $C_i$.
The LLM uses the instructions and provided relevant context to generate a short answer $a'_i$ and a long answer $l_i$ (Figure~\ref{fig:prompt3}).
Our idea is to have LLMs generate a long answer to help distil knowledge from the relevant context before concluding with a short answer.

\paragraph{Answer Normalization}
Although the short answer format has been specified, there are cases the LLM returns unexpected output, leading to lower score when assessed using exact match metrics. 
Accordingly, we have applied rules to verify those defined in the prompt.
Moreover, we observe that Wikipedia articles often use heading titles that correspond to biomedical normalization terms.
Therefore, we implement a normalization step based on Wikipedia.
For each short answer generated by the system $a'_i$, we utilize the Wikipedia API to retrieve the most relevant Wikipedia page. 
If a related Wikipedia page exists, we extract the page’s heading title to obtain the normalized answer $a_i$.
In the final hop, the answer is used as the model’s predicted output; otherwise, it is used as the anchor for the next hop.

\begin{center}
    \captionof{figure}{System prompt for answer generation}
    \label{fig:prompt3}
  \fbox{
    \begin{minipage}{0.6\textwidth}
      \scriptsize
      \texttt{<system>} \\
      You are an expert in medical and scientific knowledge. Your task is to provide accurate answers based on the provided information.  \\
      \texttt{<user>} \\
      Instructions: \\
      1. Determine the question type from the question text. \\
      2. Generate two parts in JSON format: \\
      \quad - \quad  long\_answer: a detailed explanation based on the information. \\
      \quad - \quad  short\_answer: strictly follow rules: (...) \\
      3. Do not add any extra commentary or text outside the JSON. \\      
      Information: \\      
      \quad - \quad Sub-question: \{ $Q_i$ \} \\      
      \quad - \quad  Sub-query: \{ $q_i$ \} \\      
      \quad - \quad  Anchor: \{ $a_{i-1}$ \} \\      
      \quad - \quad  Retrieval context: \{ $C_i$ \} \\    
      Response format (strict JSON): (...)
    \end{minipage}
  } \\
  \begin{minipage}{0.6\textwidth}
    \raggedleft
    {\footnotesize\itshape The text content has been condensed by substituting with (...)}
  \end{minipage}
\end{center}

\section{Experiments and Results}

\subsection{Dataset and Evaluation Metric}

We used MedHopQA dataset from Track 1: MedHopQA Track at BioCreative IX Shared Tasks~\cite{MedHopQAoverview}.
Each question focuses on disease-related topics such as signs and symptoms, genetic factors, and treatments. 
The dataset includes 45 and 1000 questions in the development set and the test set, respectively. 
To ensure test data integrity, the organizers released roughly 10,000 questions, 1,000 of which are concealed as the test set, and model performance is evaluated exclusively on those hidden questions.

We use Exact Match and Concept Level Score, the official evaluation metrics provided by the organizers from MedHopQA Track, to assess our model.
While exact match evaluates short answers at the surface level by comparing responses as strings, concept-level evaluation assesses the responses based on biomedical concepts.

\subsection{Implementation Details}
The question classification model is trained on 453 human-annotated samples, each labeled by two annotators, with conflicts resolved by a third.
It is implemented as a stacking ensemble excluding original features from the meta-learner input.
The logistic regression meta-classifier assigns weights of 0.72 and 0.28 to the outputs of the Random Forest and XGBoost models, respectively.
As a result, it achieves an accuracy of 0.89 and an F1 score of 0.83.
In the information retrieval step, the \texttt{Google Custom Search API} is used as the search engine, with a maximum of 10 results returned.
In addition, Wikipedia articles are crawled, and the sentences with the highest TF-IDF scores are selected, ensuring that the total number of tokens does not exceed 300.
Our framework integrates two LLMs: \texttt{gpt-4o-mini} for decomposing questions and generating sub-queries, and \texttt{gpt-o3-mini} for generating answers.
We set \texttt{temperature} parameter to 0 to ensure deterministic and consistent responses across models.

\subsection{Results}

In the competition, we made five submissions to experiment with and analyze the contributions of the proposed components. 
Our best result currently ranks second on the leaderboard\footnote{https://www.codabench.org/competitions/7609/}.
The results of 5 runs are shown in Table~\ref{tab:run}.
\begin{table}[!ht]
  \caption{Result of 5 runs}
  \label{tab:run}
  \begin{tabular}{cccc}
    \toprule
    \textbf{Run} & \textbf{ID} & \textbf{Exact Match Score} & \textbf{Concept Level Score}\\
    \midrule
    1 & 302553 & 0.755 &  0.812\\
    2 & 302644 & 0.783  & 0.832\\
    3 & 302784 & 0.811 & 0.836\\
    4 & 302801 & 0.829  & 0.851\\
    \textbf{5} & \textbf{302816} & \textbf{0.840} & \textbf{0.863}\\
  \bottomrule
\multicolumn{4}{r}{\small \textit{The highest results in each column are highlighted in bold}} 
\end{tabular}
\end{table}

\paragraph{Run 1: Baseline}
We evaluate the baseline model incorporating the step-by-step phases of our proposed framework. However, during the information retrieval phase, we rely solely on the snippet returned by the search engine, and no normalization is applied to the short answer.

\paragraph{Run 2: Performance improvement on yes/no questions} Since yes/no questions typically contain sufficient keywords, retrieving relevant information is relatively straightforward for this question type. 
Therefore, we initially experiment with Wikipedia-based retrieval on yes/no questions.
Specifically, this retrieval strategy is incorporated into the information retrieval phase of our baseline model. 
The improved evaluation scores indicate that the Wikipedia-based retrieval method provides useful information that helps the LLM generate more accurate answers.  
On the other hand, answers to WH-questions are taken directly from Run 1.

\paragraph{Run 3: Wikipedia-based normalization experiment}
We experiment with our proposed Wikipedia-based normalization method for short answer output. The method uses the short answers from Run 2 as input. 
The improvement in Exact Match indicates that the proposed normalization approach successfully identifies appropriate biomedical normalization terms.

\paragraph{Run 4: : Performance improvement on wh questions}
We further experiment with Wikipedia-based retrieval on WH-questions.
Specifically, this retrieval strategy is incorporated into the information retrieval phase of our model.
The Wikipedia-based normalization method is also applied for final short answer output.
On the other hand, the answers for yes/no questions are taken from Run 3.
The improved evaluation scores indicate that this method also provides useful information that helps the LLM generate more accurate answers for WH-questions.

\paragraph{Run 5: Post-processing and experiment with a stronger LLM} We reprocessed the questions that went unanswered due to processing errors and experimented with a subset of samples using a more powerful LLM, \texttt{gpt-4o-search-preview}. The table demonstrates a progressive improvement in model performance across the runs, with a peak Exact Match score of 0.840 and a Concept-Level score of 0.863 achieved in Run 5.

\section{Conclusions}
In this paper, we introduce a novel QA framework designed to efficiently handle the problem of biomedical QA. Our framework distinguishes between direct and sequential questions to apply appropriate reasoning strategies.
We propose using lightweight machine learning models to identify sequential questions before decomposition, thereby reducing unnecessary complexity and preventing errors for direct questions.
Finally, we enhance model performance by integrating information from multiple retrieval sources and leveraging in-context learning to provide rich contextual grounding. 
Our model achieved 2\textsuperscript{nd} rank on the BioCreative IX - MedHopQA Shared Task.

\section*{Acknowledgment}
We sincerely thank the organizing committee of Track 1 of the MedHopQA shared task at BioCreative IX 2025.
We also appreciate the anonymous reviewers for their insightful and valuable feedback.

\bibliography{sample-ceur}


\end{document}